\documentclass[12pt,journal,compsoc]{IEEEtran}
\usepackage{amssymb}
\usepackage{amscd}
\usepackage{amsmath}

\ifCLASSOPTIONcompsoc

\else

\fi

\ifCLASSINFOpdf

\else

\usepackage[dvips]{graphicx}
\DeclareGraphicsExtensions{.eps}
\usepackage{subfigure}
\fi

\begin{document}
%
\title{Feature Selection via Sparse Approximation for Face Recognition}
\author{Yixiong~Liang,~Lei~Wang,~Yao~Xiang, and~Beiji~Zou
\IEEEcompsocitemizethanks{\IEEEcompsocthanksitem The authors are
with the Institute of Information Science and Engineering, Central
South University, Changsha,
Hunan, 410083, China.\protect\\
E-mail: \{yxliang, wanglei, yxiang, bjzou\}@mail.csu.edu.cn
}
}

\markboth{Manuscript submitted to IEEE Trans. Pattern Anal. Mach. Intell., JULY 2010}%
{Shell \MakeLowercase{\textit{et al.}}: Bare Demo of IEEEtran.cls
for Computer Society Journals}
\IEEEcompsoctitleabstractindextext{%
\begin{abstract}
Inspired by biological vision systems, the over-complete local
features with huge cardinality are increasingly used for face
recognition during the last decades. Accordingly, feature selection
has become more and more important and plays a critical role for
face data description and recognition. In this paper, we propose a
trainable feature selection algorithm based on the regularized frame
for face recognition. By enforcing a sparsity penalty term on the
minimum squared error (MSE) criterion, we cast the feature selection
problem into a combinatorial \emph{sparse approximation} problem,
which can be solved by greedy methods or convex relaxation methods.
Moreover, based on the same frame, we propose a \emph{sparse}
Ho-Kashyap (HK) procedure to obtain simultaneously the optimal
\emph{sparse} solution and the corresponding \emph{margin} vector of
the MSE criterion. The proposed methods are used for selecting the
most informative Gabor features of face images for recognition and
the experimental results on benchmark face databases demonstrate the
effectiveness of the proposed methods.
\end{abstract}

}

\maketitle

\IEEEdisplaynotcompsoctitleabstractindextext

\IEEEpeerreviewmaketitle

\section{Introduction}\label{sec1}

\IEEEPARstart{W}{ithin} the last several decades, face recognition
has received extensive attention due to its wide range of
application from identity authentication, access control and
surveillance to human-computer interaction and numerous novel face
recognition algorithms have been proposed \cite{Zhao2003,Solar2009}.
One of the key issue to successful face recognition systems is the
development of effective face representation, namely how to extract
and select the discriminative features to represent face image.
According to the region from which features are derived, face
representation methods can be generally divided into two categories:
\emph{holistic} representation and \emph{local} representation. The
\emph{holistic} representation extract features from the whole face
image while the \emph{local} representation calculating the features
from the local faical regions.

After the introduction of the well-known \emph{Eigenfaces}
\cite{Turk1991}, the \emph{holistic} representation methods were
extensively studied
\cite{Belhumeur1997,Moghaddam1998,Bartlett2002,He2005,Wright2009,Wagner2009}.
However, local areas are often more descriptive and more appropriate
for dealing with those facial variations due to expression, partial
occlusion and illumination, since most variations in appearance only
affect a small part of the face region. Local feature analysis (LFA)
\cite{Penev1996} pioneers the study of local representation for face
recognition. Recently, local representation approaches have received
more attention and have shown more promising results
\cite{Wiskott1997,Heisele2003,Jones2003,Bicego2006,Ahonen2006,Yan2007,Albiol2008,Meyers2008,Kumar2009,Wright2009b,Cao2010}.
A lots of local feature descriptors, such as Haar-like features
\cite{Jones2003}, SIFT features \cite{Bicego2006}, histograms of
oriented gradient (HOG) \cite{Albiol2008}, edge orientation
histograms (EOH) \cite{Yan2007}, Gabor features
\cite{Wiskott1997,Lei2008}, local binary patterns (LBP)
\cite{Ahonen2006,Lei2008,Marcel2008}, Bio-inspired features
\cite{Meyers2008}, learned descriptor \cite{Cao2010} etc., have been
successfully applied in face recognition. These local features are
often over-completed, whereas only a relatively small fraction of
them is relevant to the recognition task. Thus feature selection is
a crucial and necessary step to select the most discriminant local
features to obtain a \emph{sparse} face representation. While the
prior knowledge to the choice of this local feature dictionary of
large cardinality is often limited and a consistent theory is still
missing, numerous learned methods are emerging in the empirical
practice due to their effectiveness (refer to \cite{Guyon2003} for
an excellent review of feature selection approaches in machine
learning). Adaboost-based methods are the most popular and
impressive feature selection methods in face recognition Scenario
\cite{Jones2003,Zhang2004,Yang2004,Shen2004,Shan2005,Li2007,Wang2009}.
One possible problem of these methods is very time consuming in the
training stage for the need of training and evaluating a classifier
for each feature component. An alternative is the regularized-based
method which sparsify with respect to a dictionary of features by
the sparsity-enforcing regularization techniques
\cite{Tibshirani1996,Destrero2009}. The main merits of such a
regularized approach are its effectiveness even in the presence of a
very small number of data coupled with the fact that it is supported
by well-grounded theory \cite{Destrero2009}. Another potential merit
is that the regularized methods analyze all feature components
together and may be more appropriate to capture groups of correlated
features, whereas the Adaboost-based method only consider the
relevance of each feature separately, thus may ignore the possible
dependencies between features.

Based on the regularized frame, in this paper, we propose a novel
feature selection method based on classical minimum squared error
(MSE) criterion \cite{Duda2001} which assumes a linear dependence
between the feature components and the discriminant functions. We
cast the feature selection problem into a combinatorial \emph{sparse
approximation} problem by enforcing a sparsity penalty term on the
MSE criterion and the solution can be obtained by greedy methods
such as matching pursuit (MP) \cite{Mallat1993} or the orthogonal
matching pursuit (OMP) \cite{Tropp2004,Tropp2004b,Tropp2007} and
convex relaxation methods \cite{Tropp2004b}. We restrict ourselves
to the linear models because they are relatively easy to compute and
in the absence of information suggesting otherwise, linear models
are an attractive candidates. Further, the linear model can be
extended to the nonlinear cases by explicitly or implicitly giving
some function of the local feature components. The latter is the
well-known \emph{kernel trick}.

Due to the arbitrary selection of \emph{margin} vector, the MSE
procedure cannot guarantee to obtain the optimal separating vector
even in the separable case \cite{Duda2001}. We impose the
\emph{sparse} constrains on the Ho-Kashyap (HK) procedure
\cite{Duda2001} and propose a named \emph{sparse} HK (SHK) procedure
to obtain simultaneously the optimal \emph{sparse} solution and the
corresponding \emph{margin} vector. Similar to the original HK
procedure, the proposed SHK procedure is an iterative scheme that
alternates between solution of the \emph{sparse} vector based on the
current \emph{margin} vector and a process of updating the
\emph{margin} vector. It is flexible and can work with any greedy
methods or convex relaxation methods.

Gabor and LBP are two most representative local features in face
recognition. We select Gabor feature as the start representation due
to its peculiar ability to model the spatial summation properties of
the receptive fields of the so called "bar cells" in the primary
visual cortex. Then we apply the proposed feature selection method
to select the most informative Gabor features for face recognition.
Experimental results on the benchmark face databases demonstrate the
effectiveness of the proposed feature selection methods.

Our method may be mainly inspired by the work \cite{Destrero2009}
which is also applying the \emph{sparse} regularized term to the
linear model to perform the feature selection. Nevertheless, the
linear model in \cite{Destrero2009} neglects the \emph{bias} on the
one hand and only enforces the linear dependence between the feature
components and the class labels on the other hand. In fact, this
simple linear dependence is equivalent to set all entries of the
\emph{margin} vector equal to 1 in MSE criterion model. Our method
starts off with the MSE criterion and considers simultaneously the
\emph{bias} and the adaptive \emph{margin} vector and hence can be
seemed as a generalization of the method in \cite{Destrero2009}.
Moreover, in \cite{Destrero2009} the \emph{sparse} solution is
obtained through iterative soft-thresholding method and the
convergence relies on the careful normalization of each features
component of all training samples at a time, which may destroy the
structure of the features. Our method adheres to the original
features without any additional normalization and also obtain the
convergence.

The rest of this paper is organized as follows. In Section
\ref{sec2}, we start off with the MSE criterion and propose a novel
feature selection method based on sparsity-enforcing regularized
techniques. Based on the same frame, in Section \ref{sec3}, we
present a \emph{sparse} extension of the classical HK procedure for
feature selection. In Section \ref{sec4} we first briefly review the
Gabor face representation and then illustrate how to apply the
proposed feature selection frame to select the most informative
Gabor features for face recognition. Experiments and analysis are
described in Section \ref{sec5}, whereas Section \ref{sec6}
concludes the paper.

\section{Feature selection based on \emph{sparse} MSE Criterion} \label{sec2}
In this section, we present a new feature selection algorithm based
on the MSE criterion. As mentioned before, we restrict ourselves to
the case of a linear discriminant functions that are linear in the
components of feature $\mathbf{x}=[x_{1},\ldots,x_{d}]^{T}$:
\begin{equation}\label{eq1}
    g(\mathbf{x})=\sum_{i=1}^{d}\omega_{i}x_{i}+\omega_{0}=\mathbf{y}^{T}\mathbf{a},
\end{equation}
where $g(\mathbf{x})$ denotes the discriminant function;
$\omega_{0}$ is the \emph{bias} or \emph{threshold};
$\omega_{i}(i=1,\dots,d)$ is the weights;
$\mathbf{y}=[1,x_{1},\ldots,x_{d}]^{T}$ and $\mathbf{a} =
[\omega_{0},\dots,\omega_{d}]^{T}$ are the augmented feature vector
and augmented weight vector, respectively. Since the face
recognition can be cast into a classification of the intra-personal
and extra-personal variation \cite{Moghaddam1998}, we focus on a
binary classification problem. As suggested in \cite{Duda2001}, we
substitute all negative samples (i.e. extra-personal variations)
with their negatives to forget the labels and look for a weight
vector such that $\mathbf{y}_{i}^{T}\mathbf{a}>0$ for all of the
samples. Indeed, this relation is invariant under a positive scaling
of $\mathbf{a}$. Thus, we can define a canonical hyperplane such
that $\mathbf{y}_{i}^{T}\mathbf{a}=b_{i}$ where $b_{i}$ is a
positive constant called the \emph{margin}. Now the problem can be
reformulated as the following linear system of equations:
\begin{equation}\label{eq2}
    \mathbf{Y}\mathbf{a}=\mathbf{b},
\end{equation}
where $\mathbf{Y}=[\mathbf{y}_{1},\dots, \mathbf{y}_{n}]^{T}\in
\mathcal{R}^{n\times (d+1)}$ is the augmented feature matrix and
$b=[b_{1},\dots,b_{n}]^{T}$ is the \emph{margin} vector. Due to the
size of $\mathbf{Y}$, it is infeasible to obtain the exact solution
of (\ref{eq2}). One classical relaxation is to solve the minimize
squared error criterion function
\begin{equation}\label{eq3}
    \min\|\mathbf{Y}\mathbf{a}-\mathbf{b}\|_{2}^{2}=\min\sum_{i=1}^{n}(\mathbf{y}_{i}^{T}\mathbf{a}-b_{i})^{2}.
\end{equation}
It can be solved by a gradient search procedure. However, the MSE
solution do not provide feature selection in the sense because it's
typically \emph{non-sparse}. By enforcing \emph{sparse}
regularization term on the MSE criterion, we can turn the feature
selection into solving the following sparsity-enforcing MSE (SMSE)
criterion:
\begin{equation}\label{eq4}
    \min
    \|\mathbf{Y}\mathbf{a}-\mathbf{b}\|_{2}^{2}+\tau^{2}\|\mathbf{a}\|_{0},
\end{equation}
where $\|\cdot\|_{0}$ is the $l_{0}$ \emph{quasi-norm} counting the
nonzero entries of a vector and $\tau$ is a threshold that
quantifies how much improvement in the approximation error is
necessary before we admit an additional term into the approximation.
It is a classic combinatorial \emph{sparse approximation} problem
and can be solved by greedy techniques such as MP and OMP which
construct a \emph{sparse} approximant one step at a time by
selecting the atom most strongly correlated with the residual part
of the signal and use it to update the current approximation. An
alternative to solving the SMSE criterion (\ref{eq4}) is the convex
relaxation methods which replace the problem with a relaxed version
that can be solved more efficiently. The $l_{1}$ \emph{norm}
provides a natural convex relaxation of the $l_{0}$
\emph{quasi-norm}, and it suggests that we may be able to solve
\emph{sparse approximation} problems by introducing an $l_{1}$
\emph{norm} in place of the $l_{0}$ \emph{quasi-norm}. From this
heuristic, a relaxed version of SMSE (RSMSE) criterion can be
derived as follows:
\begin{equation}\label{eq5}
    \min
    \frac{1}{2}\|\mathbf{Y}\mathbf{a}-\mathbf{b}\|_{2}^{2}+\gamma\|\mathbf{a}\|_{1},
\end{equation}
which is an unconstrained convex function and thus standard
mathematical programming softwares can be used to find a minimizer.
The parameter $\gamma$ negotiates a compromise between approximation
error and sparsity. It has been proved that if the feature matrix
$\mathbf{Y}$ is incoherent and the threshold parameters are
correctly chosen, then the solution to RSMSE criterion (\ref{eq5})
identifies every significant atom from the solution to SMSE
criterion (\ref{eq4}) and no others \cite{Tropp2004b}.

From a run-time point of view, we adopt OMP to solve the SMSE
criterion (\ref{eq4}). Since OMP is iterative, we must supply a
criterion for stopping the iteration. Here are two possibilities:
\begin{itemize}
  \item One may halt the procedure when the norm of the residual declines
below a specified threshold.
  \item One may halt the procedure after pre-defined number of distinct feature
  components have been selected.
\end{itemize}
In our implementation, the iteration is stopped whenever one of the
above conditions is satisfied.

Notice that in the criterion (\ref{eq3}) (\ref{eq4}) and
(\ref{eq5}), the entries of the \emph{margin} vector $\mathbf{b}$
are arbitrary positive constants. Obviously, different choice of
$\mathbf{b}$ would typically lead to different solutions. As the MSE
solution is directly related to the Fisher discriminant vector with
a proper choice of the \emph{margin} vector (i.e. the entries
$b_{i}$ corresponding to the same class are equal to the ratio of
the sample size of this class to the total sample size)
\cite{Duda2001}, the SMSE solution or RSMSE solution gives a natural
\emph{sparse} generalization of Fisher linear discriminant.
Hereafter we refer to the resulting feature selection algorithm as
\emph{sparse} Fisher (SFisher) procedure. Moreover, if we set
$\mathbf{b} = \mathbf{1}_{n}$ and $\omega_{0}=0$ (we refer to the
resulting algorithm as simplified SMES procedure, or SSMES), in this
special case the RSMSE criterion (\ref{eq5}) degenerates into the
linear model described in \cite{Destrero2009}, and thus our method
can also be seemed as a generalization of the method in
\cite{Destrero2009}.

\section{Feature selection based on \emph{sparse} HK procedure}\label{sec3}
Because the objective is minimizing
$\|\mathbf{Y}\mathbf{a}-\mathbf{b}\|_{2}^{2}$, as discussed in
\cite{Duda2001}, the MSE procedures yield a solution whether the
samples are linearly separable or not, but there is no guarantee
that this vector is a separating vector even in the separable case.
However, in the separable case, there do exist a \emph{margin}
vector $\mathbf{\hat{b}}$ with all positive entries such that the
corresponding MSE solution is the separating vector. The HK
procedure extends the MSE procedure to deal with this problem by
determining $\mathbf{a}$ and $\mathbf{b}$  alternately where the
components of $\mathbf{b}$  cannot
decrease. 
Borrowing from the same ideas, in this section we propose a
\emph{sparse} version the HK procedure to extend our method
described in the former section. Specifically speaking, in the
proposed SHK procedure there are two stages for each iteration: one
for \emph{sparse} approximating that essentially evaluates
$\mathbf{a}$ and one for updating the \emph{margin} vector
$\mathbf{b}$. \emph{Sparse} approximating can be conveniently
performed by using greedy or convex relaxation algorithms to solve
the SMES criterion (\ref{eq4}) with a given $\mathbf{b}$. Similar
original HK procedure, the updating rule of $\mathbf{b}$ is to start
with $\mathbf{b}>0$\footnote{$\mathbf{b}>0$ means that every
component of $\mathbf{b}$ is positive.} and to refuse to reduce any
of its components, namely
\begin{equation} \label{eq6}
\left\{ \begin{aligned}
         \mathbf{b}(1)&> 0 \\
         \mathbf{b}(t+1)&= \mathbf{b}(t)+2\eta(t)\mathbf{e}^{+}(t),
\end{aligned} \right.
\end{equation}
where $\eta(t)$ is a positive scale factor or \emph{learn rate};
$\mathbf{e}(t)=\mathbf{Y}\mathbf{a}(t)-\mathbf{b}(t)$ is the error
vector and
$\mathbf{e}^{+}(t)=\frac{1}{2}(\mathbf{e}(t)+|\mathbf{e}(t)|)$ is
the positive part of the error vector, respectively. Given some
stopping rules, our algorithm is:\\ \textbf{Algorithm
SHK.}\\Initialization: Set $\mathbf{b}(0)>0$, $0<\eta(\cdot) < 1$.
Set the iteration index $t=1$.
\\Repreat until convergence (stopping criterion):
\begin{itemize}
  \item \emph{Sparse approximation} stage: Use any greedy algorithms or
  convex relaxation methods to computer $\mathbf{a}(k)$ by approximating the solution of SMES
  criterion (\ref{eq4}).
  \item \emph{Margin} vector update stage: $\mathbf{b}(t+1)=
  \mathbf{b}(t)+2\eta(k)\mathbf{e}^{+}(t)$.
  \item Set $t=t+1$.
\end{itemize}
In our implementation, the stopping rule is that when
$\|\mathbf{b}(t+1)-\mathbf{b}(t)\|_{2}<\epsilon$ is reached, the
loop is terminated.

It is noteworthy that although the convergence of the original HK
procedure can be proven theoretically \cite{Duda2001}, owing to the
introduction of the \emph{sparse approximation} stage, exact
analysis of the convergence of the proposed SHK algorithm in a
deterministic manner is rather complicated or even impossible.
Nevertheless, we can obtain the convergence by careful selection of
$\eta(t)$ which decreases with $t$. Our choice is to set
$\eta(t)=\frac{\eta(1)}{t}$.

\section{Gabor feature selection for face recognition}\label{sec4}
In this section we describe how we specialize the proposed feature
selection frame to the case of face recognition. We first briefly
review the Gabor representation of face and then describe how to
apply the proposed feature selection methods to select Gabor
features for face recognition.
\subsection{Gabor representation}
We start with the widely used Gabor representation because the
kernels of Gabor filters are similar to the 2D receptive field
profiles of the mammalian cortical simple cells and exhibit
desirable characteristics of spatial locality and orientation
selectivity \cite{Wiskott1997,Yang2004,Shan2005}. The Gabor
representation of a face image can be obtained by convolving the
image by a set of Gabor filters which are commonly defined as
follows:
\begin{equation}\label{eq7}
    \psi_{\mu,\nu}=\frac{\|\mathbf{k}_{\mu,\nu}\|_{2}^{2}}{\sigma^{2}}e^{-\frac{\|\mathbf{k}_{\mu,\nu}\|_{2}^{2}\|\mathbf{z}\|_{2}^{2}}{2\sigma^{2}}}[e^{i\mathbf{k}_{\mu,\nu}\mathbf{z}}-e^{-\frac{\sigma^{2}}{2}}],
\end{equation}
where $\mathbf{z}$ is the coordinate vector; parameters $\mu$ and
$\nu $ define the orientation and the scale of the Gabor filter;
parameters $\sigma$ is the standard deviation of Gaussian window;
$\mathbf{k}_{\mu,\nu}$ is the wave vector given by
$\mathbf{k}_{\mu,\nu}=k_{\nu}e^{i\phi_{\mu}}$, where
$k_{\nu}=\frac{k_{max}}{f^{\nu}}$ and $\phi_{\mu}=\frac{\pi\mu}{8}$
if eight different orientations have been chosen; $k_{max}$ is the
maximum frequency, and $f$ is the spatial factor between kernels in
the frequency domain. In face recognition area, researchers commonly
use 40 Gabor filters with five scales $\nu\in\{0,\cdots,4\}$ and
eight orientations $\mu\in\{0,\cdots,7\}$ and with $\sigma=2\pi$,
$k_{max}=\frac{\pi}{2}$ and $f=\sqrt{2}$. However, we set the scale
ranges from -1 to 2 rather than from 0 to 4 due to the using of
smaller size of face images in our experiments. Thus only 32 Gabor
filters are used. Convolving the face image with these 32 Gabor
filters and only extracting the magnitudes information can then
generate a high dimensional Gabor representation. For example, for
an image with $64\times64$ pixels, the total number of Gabor
features is $4\times8\times64\times64=131,072$.

A noticeable problem in discrete convolution is the choice of proper
size of the convolution mask. It should be large enough to show the
nature of Gabor kernels and not be too large for the computation
efficiency. As suggested in \cite{Dunn1995}, we truncate the Gabor
filters to six times the span of the Gaussian function. As the span
of Gaussian function is $\frac{\sigma}{k_{\nu}}$, the Gabor mask is
then truncated to a width
$w=\frac{6\sigma}{k_{\nu}}+1=24\times2^{\frac{\nu}{2}}+1$. Thus in
our experiments the size of Gabor filters are $19\times19$,
$25\times25$, $35\times35$, $49\times49$ corresponding to the scale
of $\nu\in\{-1,\cdots,2\}$.

\subsection{Feature selection for face recognition}
Now it time to turn our attention to the feature selection of the
high dimensional Gabor representation. Similarly to Moghaddam
\emph{et al.} \cite{Moghaddam1998}, we temporarily cast the face
recognition into a classification of the intra-personal (hereafter
as positive) and extra-personal (hereafter as negative) variation.
For each pair of face images $I_{i}$ and $I_{i}'$, we compare the
corresponding Gabor feature components. Specifically, for each pair
of input images we obtain a feature vector $\mathbf{x}_{i}$ whose
elements are the absolute difference between the corresponding Gabor
representations. Given a training set, we can then get the augmented
feature matrix $\mathbf{Y}$ following the routine described in
Section \ref{sec2}.

A by-no-means negligible problem in practical is the overwhelmingly
huge size and unbalance of the training samples \cite{Jones2003}.
For instance, given a training set that includes $K$ images for each
of the $C$ individuals, the total number of image pairs is $CK
\choose 2$ whereas only a small minority, $C$$K \choose 2$ of these
pairs display the intra-personal variation. Let $C=300$ and $K=4$,
then the size of positive samples and negative samples are $1,800$
and $717,600$ respectively with their ratio be close to $1:400$.

Obviously, such a huge samples size will lead to severe memory and
computational problem. In addition, the unbalance training samples
may bias the performance of the feature selection. In order to
obtain balanced systems of reasonable size, we randomly sample the
positive and negative samples with a comparable ratio to build the
augmented feature matrix $\mathbf{Y}$. In practical, we can sample
negative samples while keeping all positive samples with their ratio
varying from $1:1$ to $1:10$.

Once we build the augmented feature matrix $\mathbf{Y}$, we can find
the solution of SMSE criterion (\ref{eq4}) with a given or an
adaptive \emph{margin} vector $\mathbf{b}$ according to the
procedure previously described. Then the Gabor feature components
corresponding to non-zero entries of the augmented weight vector
$\mathbf{a}$ are selected as the most informative ones and used for
further face recognition.

Recalled that the above feature selection frame based on linear
discriminant functions also establishes a linear classifier with a
\emph{bias} $w_{0}$ discriminating the intra-personal and
extra-personal difference, so it can be used for face recognition
directly. However, one can also consider its usage as a pure feature
selection tool to reduce the numbers of Gabor features and adopt
some other common classifiers such as \emph{nearest neighbor
classifier} (NNC), \emph{Fisher classifier} (FC) \cite{Duda2001} or
\emph{support vector machines} (SVM) \cite{Chang2001} for the
recognition.

\section{Experiments and results} \label{sec5}

In order to evaluate the proposed approach, we carry out some
experiments on two large face databases: CAS-PEAL-R1 \cite{Gao2008}
and LFW \cite{Huang2007} face database. The CAS-PEAL-R1 face
database contains $30,863$ images of $1,040$ Chinese subjects with
different variations of pose, expression, accessories, age, and
lighting. The LFW face database contains $13,233$ labeled face
images collected from news sites in the Internet. These images
belong to $5,749$ different individuals and have high variations in
position, pose, lighting, background, camera and quality. Therefore
LFW database is more appropriate to evaluating face recognition
methods in realistic and unconstrained environments.

In all our experiments, each image is rotated and scaled so that the
centers of the eyes are placed on specific pixels and then was
cropped to $64\times64$ pixels \footnote{The eyes locations are
given in CAS-PEAL-R1 database. For LFW database, we adopted standard
fiducial point detector to extract the eyes locations and annotated
them manually whenever the automatic eyes locator failed.}. As
described before, we only select 32 Gabor filters to extract the
Gabor features.

\subsection{Results on CAS-PEAL-R1 database}
We restrictively follow the CAS-PEAL-R1 evaluation protocol which
specifies one training set, one gallery set and six probe sets
\cite{Gao2008}. Therefore the training sets include $1,200$ images
of 300 subjects and the ratio of intra-personal sample size to
extra-personal sample size is $1,800:717,600$. We keep all
intra-personal samples while randomly sampling the extra-personal
samples with a ratio of $1:7$. If all Gabor features are considered,
the linear problem we are about to build is rather large. In fact,
the size of the augmented feature matrix $\mathbf{Y}$ is come to
$131,073\times14,400$. Obviously direct multiplication on such a
matrix is infeasible. One possible choice is to reduce the number of
Gabor features if possible. With the prior knowledge of that the
magnitude of the Gabor filters is not sensitive to the positions, we
can reduce the number of positions by a simply down sampling scheme
with a factor 16. Thus the number of positions is roughly one
sixteenth of the total number of pixels. So after the down sampling,
the size of the augmented feature matrix $\mathbf{Y}$ is reduced to
$8193\times14,400$.
\subsubsection{Feature selection results}
We conducted experiments on the CAS-PEAL-R1 training set using
SSMES, SFisher and SHK procedure to select 500 most informative
Gabor features, respectively. Their characteristics can be observed
by their statistics. The location distribution of selected Gabor
\begin{figure}[!h]
\centering \includegraphics[width=0.5\textwidth]{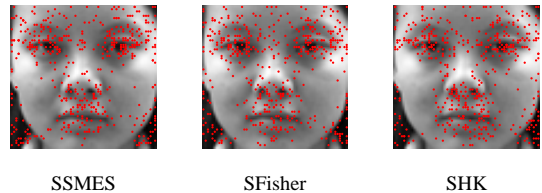}
\caption{Location distribution of the first 500 selected Gabor
features on CAS-PEAL-R1 dataset.} \label{fig_LocDistribution}
\end{figure}
features are shown in Fig. \ref{fig_LocDistribution}. It is
interesting to see that most of selected Gabor features resulting
from all three methods are located around the prominet facial
features such as eyebrows, eyes, nose and mouth, while seldom being
located on the cheek area. This indicates that the prominent facial
features regions carry the most important discriminating information
while the cheek region conveying less information.  Moreover, a
minority of selected features are located on external features such
as cheek contour and jaw line. In fact, although the external region
does not cover the face much, the external features implicitly uses
shape information and thus are useful for distinguishing thin faces
from round faces. This result is agreed with Ref. \cite{Zou2007}.
\begin{figure}[!h]
\centering \subfigure[ ]{ \label{fig:subfig:kerDistribution1}
\includegraphics[width=0.35\textwidth]{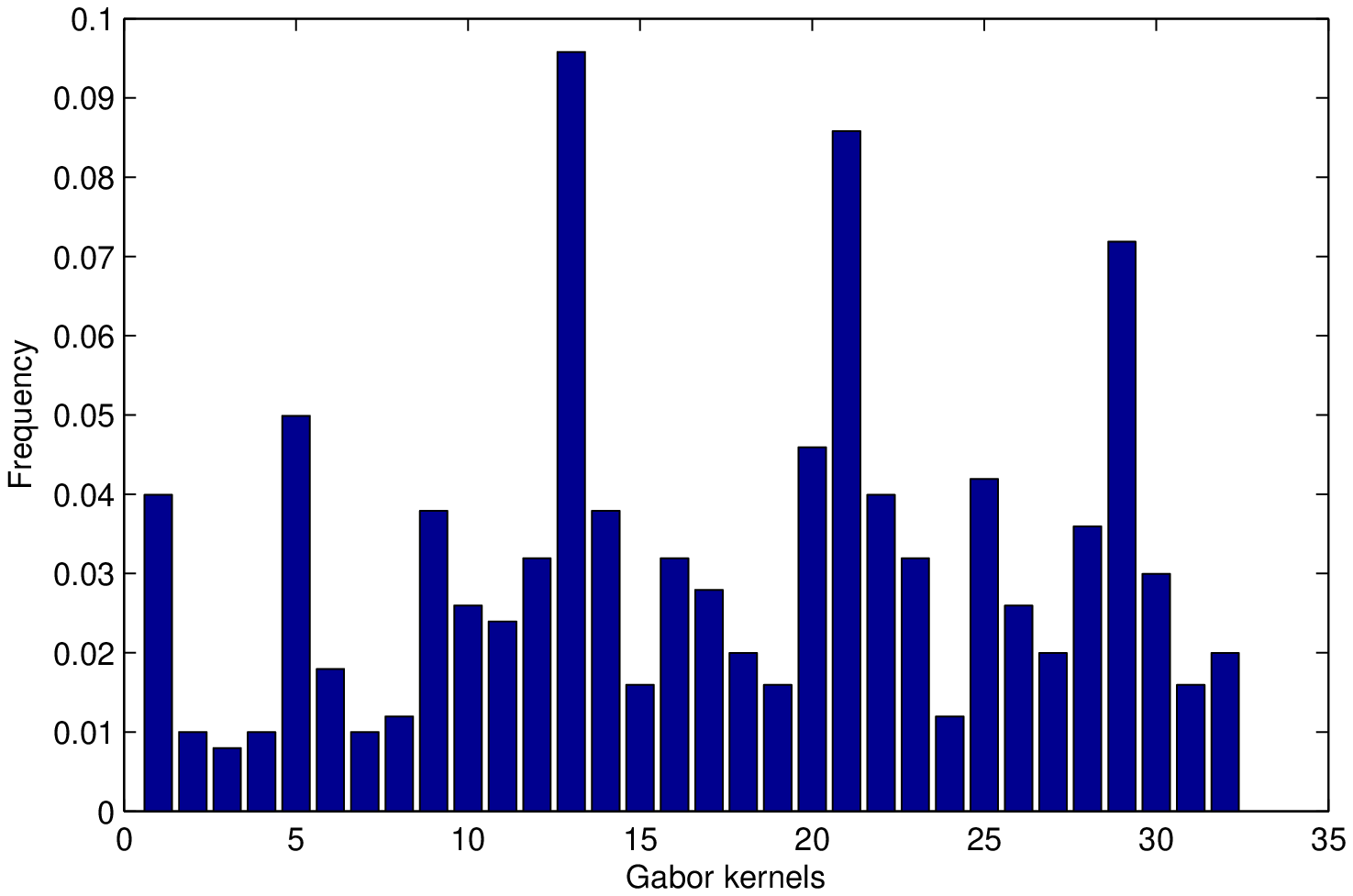}}
\subfigure[ ]{ \label{fig:subfig:kerDistribution2}
\includegraphics[width=0.35\textwidth]{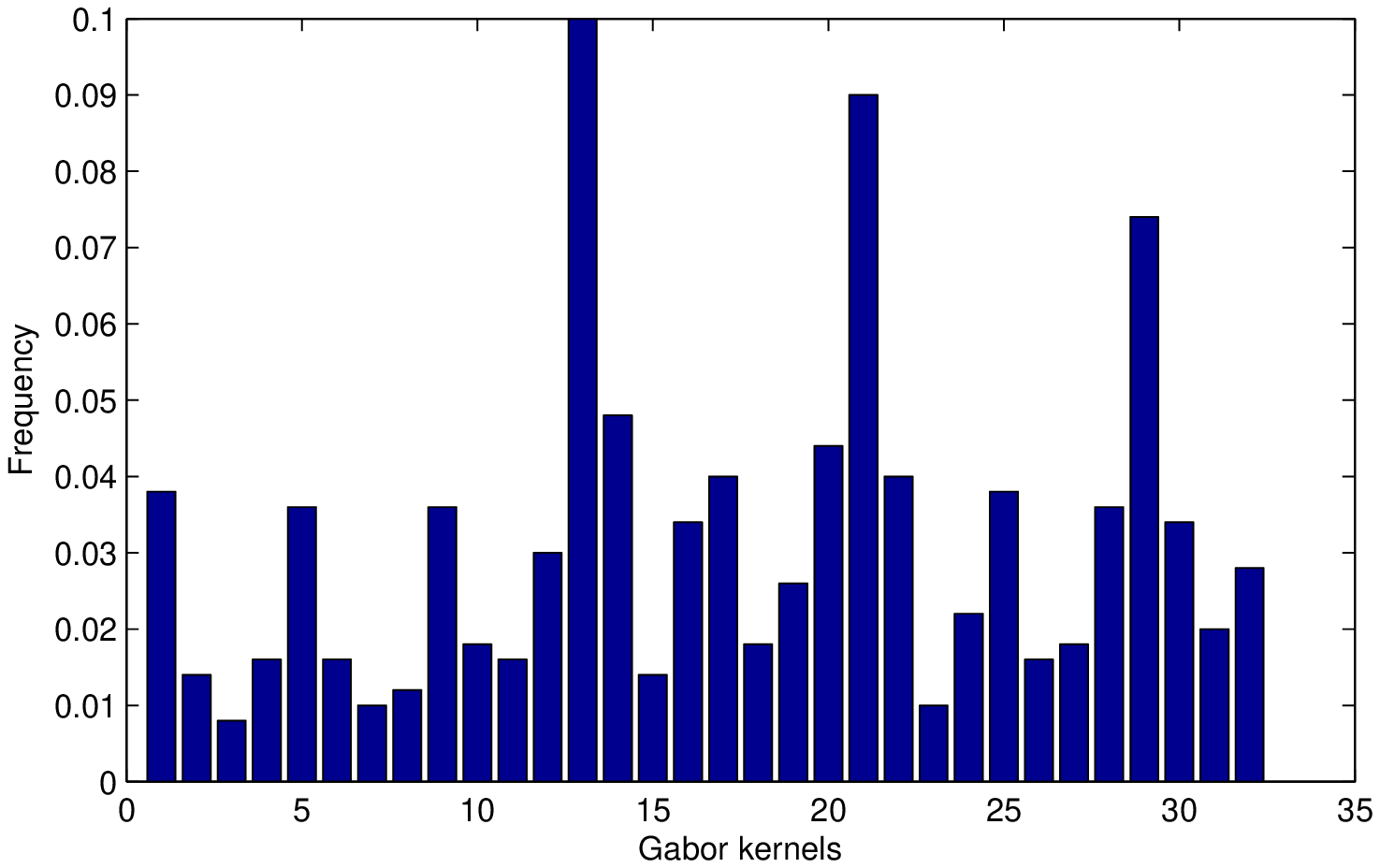}}
\subfigure[ ]{ \label{fig:subfig:kerDistribution3}
\includegraphics[width=0.35\textwidth]{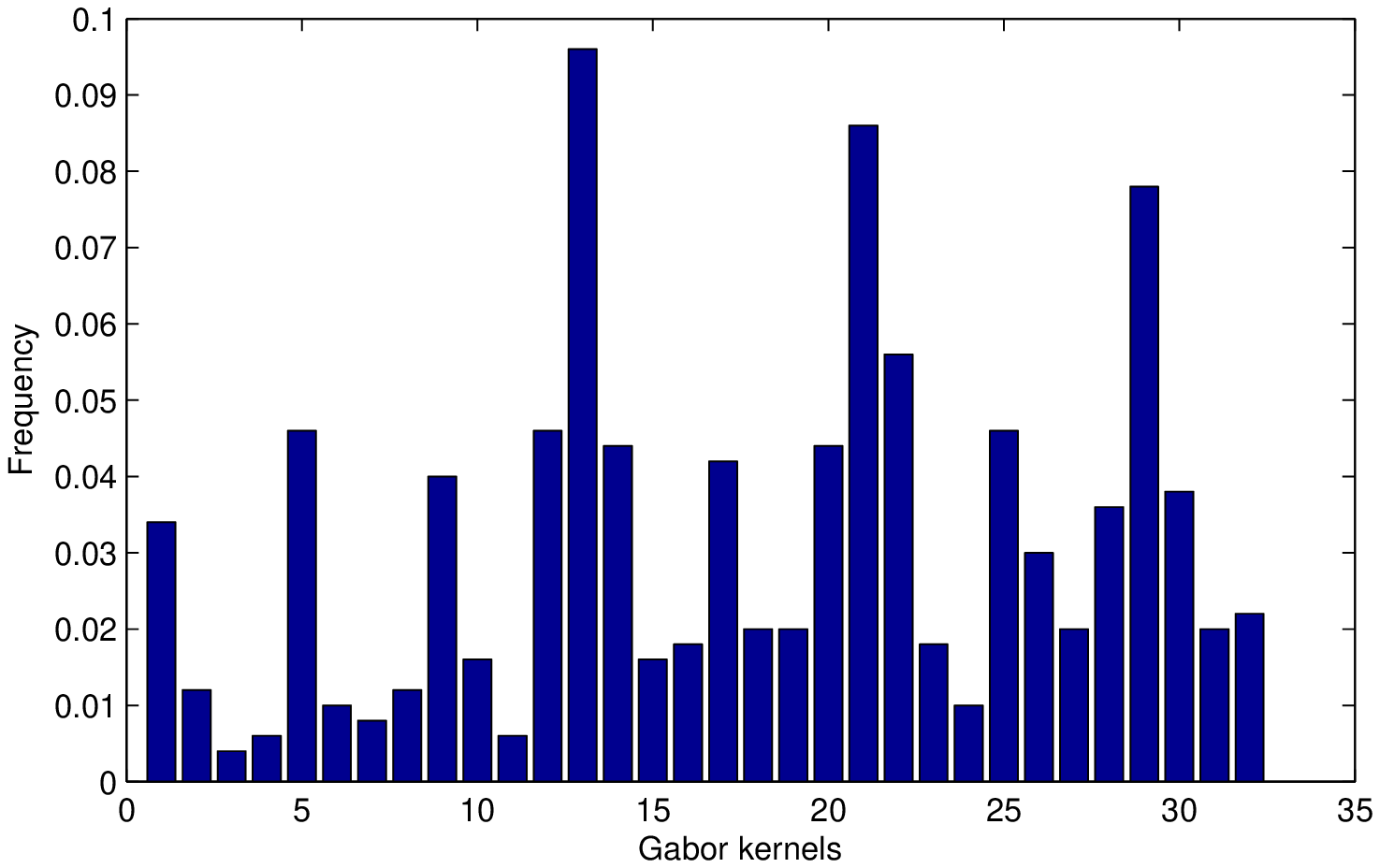}}
\caption{Distribution of 32 Gabor kernels in the first 500 leading
Gabor features on CAS-PEAL-R1 dataset. (a) SSMES. (b)
SFisher.(c)SHK.} \label{kerDistribution}
\end{figure}

We also compared the frequency of Gabor kernels in the selected
Gabor features. Fig. \ref{kerDistribution} illustrates the frequency
of the 32 Gabor kernels in the leading 500 Gabor features selected
by SSMES, SFisher and SHK procedure. Obviously, different scales and
orientations contribute different and the distribution of features
selected by different methods is somewhat uniform: the 0-scale and
1-scale are more likely important than the other two scales and most
of horizontal and vertical Gabor kernels have extracted stronger
features than those with other orientation.

\subsubsection{Classification results}
The selected Gabor features are then adopted for face recognition.
The classical classifiers, NNC and FC, are chosen to recognize the
faces. As mentioned above, the proposed feature selection frame can
perform the intra-personal and extra-personal recognition task. Thus
we also used it for face recognition by treating the face
recognition as a series of \emph{pair matching} problems. However,
in many situations there are more than one subject satisfying the
separating condition. In order to make a final decision we simply
classify the unknown face as the subject whose samples can maximize
the linear discriminant function (\ref{eq1}), i.e. the
\emph{margin}. Therefore in some sense it can be seen as a maximum
\emph{margin} classifier (MMC).

We also implemented 3 previous Gabor-based approaches for
comparison. The first is using Gabor feature without feature
selection for face representation and NNC for recognition, which is
denoted as "G+NNC". The second method "G+FC" denotes the GFC method
in \cite{Liu2002}, i.e. the PCA+LDA on down-sampled Gabor features.
The third method, "G Ada+FC", is the AGFC method in \cite{Shan2005}
which using Adaboost to select Gabor features and FC for
classification. For clarity, "G SSMES+NNC", "G SFisher+NNC" and "G
SHK+NNC" respectively denote the method using SSMES, SFisher and SHK
procedure to select Gabor features and NNC for recognition.
Similarly, for the other two classifiers, the corresponding methods
are denoted as "G SSMES+MMC", "G SFisher+MMC", "G SHK+MMC" and "G
SSMES+FC", "G SFisher+FC", "G SHK+FC". We investigated 3 kinds of
distance measurements: $l_{1}$ distance, $l_{2}$ distance and
\emph{cosine} distance, and found that for NNC, $l_{1}$ distance
achieves the best performance while for FC, the \emph{cosine}
distance performing best. Thus we selected $l_{1}$ distance for NNC
and \emph{cosine} distance for FC. In our implementation, the number
of Gabor features used in "G+FC" is down-sampled to the dimension of
$8,192$ and in "G Ada+FC", $2,000$ Gabor features are selected by
Adaboost. The optimal dimension for PCA and LDA are determined by
testing all possible dimensions. The results on 5 different probes
sets are shown and compared in Table \ref{tab1}.

\begin{table}[h]
\caption{Recognition Performance comparisons on different
CAS-PEAL-R1 probe sets (\%)}\label{tab1}
\begin{tabular}{|l|c|c|c|c|c|}
  \hline
  Methods & Age & Exp & Dis & Bac & Acc \\
  \hline
  G+NNC &78.79&81.34&98.18&93.49&63.63 \\
  G+FC &96.97&\textbf{92.68}&\textbf{98.91}&98.19&\textbf{84.55} \\
  G SSMES+NNC &87.88&78.79&95.64&94.94&68.32 \\
  G SSMES+MMC &7.58&8.47&17.45&19.71&5.38 \\
  G SSMES+FC &\textbf{100.00}& 89.94&98.18&97.47&79.34 \\
  G SFisher+NNC &93.94&77.26&96.00&94.94&68.45\\
  G SFisher+MMC &74.24&74.97&93.82&91.68&49.85 \\
  G SFisher+FC &\textbf{100.00}&90.32&97.82&97.65&80.88 \\
  G SHK+NNC &93.94&79.87&96.36&95.12&69.98 \\
  G SHK+MMC &75.76&77.01&94.54&92.22&52.12 \\
  G SHK+FC &\textbf{100.00}&91.34&\textbf{98.91}&\textbf{98.92}&80.96 \\
  G Ada+FC &96.97&89.81&96.73&97.83&78.77 \\
  \hline
\end{tabular}
\end{table}

From Table \ref{tab1}, we can obtain several major observations.
First, although the proposed feature selection frame also
establishes a classifier which can be straightforwardly used for
face recognition, its performances are  not as satisfactory as
expected, especially for the "G SSMES + MMC" method. Our explanation
is that though the feature selection frame can select effectively
the meaningful features, it may overestimate or underestimate the
corresponding weights, leading to the \emph{over-fitting} problems.
Comparing to the "G SSMES + MMC", the classifiers used in "G SFisher
+ MMC" and "G SHK + MMC" both consider the \emph{bias} and the
\emph{margin} and thus achieve better results. The second
observation is that the FC based methods ("G+FC", "C SMESS+FC", "C
SFisher+FC", "C SHK+FC" and "G Ada+FC") perform much better than the
other two classifiers based methods. In general, the algorithms with
regularized-based feature selection procedure only use 500 Gabor
features and slightly outperform "G Ada+FC" with $2,000$ Gabor
features selected by Adaboost and is comparable to "G+FC" using
$8,192$ Gabor features, which shows that the proposed feature
selection frame is effective for face recognition. Third, SFisher
and SHK perform better than SSMES in the sense of both feature
selection and classification. This results indicate that the
consideration of \emph{bias} and the \emph{margin} will not make the
learning process overfits the training data but increase the
generalizability.

\subsection{Results on LFW database}
We also conducted some experiments on the LFW database for further
investigation. Unlike the CAS-PEAL-R1 database, the LFW database
have larger degree of variability and the recognition is only to be
done by \emph{pairs matching}, instead of searching for the most
similar face in the database. We still followed their protocol which
gives two \emph{Views}: \emph{View} 1 for model selection and
algorithm development while \emph{View} 2 for performance reporting.
\emph{View} 1 specifies one training set containing $2,200$ pairs
and one testing set containing $1,000$ pairs. \emph{View} 2 consists
of ten sets with 600 images in each case. They can be combined into
10 different training/testing set pairs. In our experiments, the
training set of \emph{View} 1 are chosen for training the feature
selection model and the performance are reported using 10-fold cross
validation on the \emph{View} 2.

\begin{figure}[!h]
\centering \includegraphics[width=0.45\textwidth]{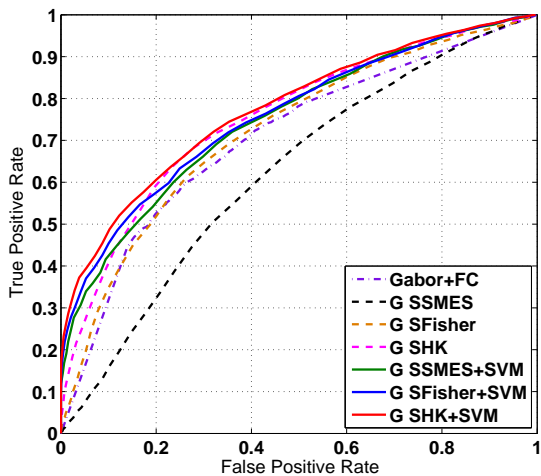}
\caption{ROC curve comparison on \emph{View} 2 of the LFW data set.
Each point on the curve represents the average over the 10 folds of
(false positive rate, true positive rate) for a fixed threshold. }
\label{fig_ROC}
\end{figure}

The proposed feature selection frame is used as a feature selector
to select 500 most informative Gabor features from the original
$131,072$ original features. We directly adopted the proposed frame
as a classifier to recognize the unknown \emph{pairs} in company
with the SVM classifier. The corresponding methods are referred as
"G SMESS", "G SFisher", "G SHK" and "G SMESS+SVM", "G SFisher+SVM",
"G SHK+SVM" respectively. We also investigated the performance of
the method "G+FC" which uses all $131,072$ original features as
representation and FC as a classifier. The results of the
experiments are described in Table \ref{tab2} below and the ROC
comparison curves of different methods are illustrated in Fig.
\ref{fig_ROC}.\begin{table}[h] \caption{Mean ($\pm$ standard error)
recognition accuracy on \emph{View} 2 of LFW data set
(\%)}\label{tab2} \centering
\begin{tabular}{|l|c|}
  \hline
  Methods & Recognition accuracy \\
    \hline
G+FC & 67.10 $\pm$ 0.53 \\
G SSMES & 60.60 $\pm$ 0.64 \\
G SFisher & 66.70 $\pm$ 0.49 \\
G SHK &68.27 $\pm$ 0.58 \\
G SSMES+SVM & 68.30 $\pm$ 0.59 \\
G SFisher+SVM & 69.18 $\pm$ 0.52 \\
G SHK+SVM &\textbf{70.32 $\pm$ 0.44} \\
  \hline
\end{tabular}
\end{table}

As can be seen, a direct application of proposed feature selection
frame as a classifier perform somewhat worse than the performance
achieved by using a SVM classifier. Recalled that the "G SFisher"
algorithm actually performs a \emph{sparse} Fisher classification
which only uses 500 features and achieves a comparable performance
of the "G+FC" method using $131,072$ original features both in terms
of accuracy and ROC curve (the accuracy is slightly lower, but the
ROC performance is better). This phenomena further demonstrates the
effectiveness of the proposed feature selection frame. Again,
SFisher and especially SHK perform better than SSMES in the sense of
both feature selection and classification in this dataset, which can
be attributed to the consideration of the \emph{bias} and adaptive
\emph{margin} in the linear model (\ref{eq2}).

\section{Conclusion} \label{sec6}
We have presented a novel feature selection algorithm based on
well-grounded sparsity-enforcing regularization techniques for face
recognition. We cast the feature selection problem into a
combinatorial \emph{sparse} \emph{approximation} problem by
enforcing a sparsity penalty term on the MSE criterion, which can be
solved by greedy methods or convex relaxation methods. Moreover, we
introduced the sparsity constrain into the traditional HK procedure
and proposed a \emph{sparse} HK procedure to obtain simultaneously
the optimal \emph{sparse} solution and the corresponding
\emph{margin} vector of the MSE criterion. The proposed frame was
applied to select most informative Gabor features for face
recognition and the experimental results on CAS-PEAL-R1 face
database and LFW face database are favorable to the previous
state-of-the-art Gabor-based methods. Our future work includes
exploring other more effective low-level face representation and
other sophisticated classification strategy to produce better
performance.



\ifCLASSOPTIONcompsoc
  \section*{Acknowledgments}
\else
  \section*{Acknowledgment}
\fi

This research is partially supported by the National Natural Science
Funds of China (No. 60803024 and No. 60970098), the Doctoral Program
Foundation of Institutions of Higher Education of China (No.
200805331107 and No. 20090162110055), the Major Program of National
Natural Science Foundation of China (No. 90715043), and the Open
Project Program of the State Key Lab of CAD\&CG (No. A0911 and No.
A1011 ), Zhejiang University.

\ifCLASSOPTIONcaptionsoff
  \newpage
\fi

\end{document}